%% file: 0_acl_latex.tex
\pdfoutput=1

\documentclass[11pt]{article}

\usepackage[final]{acl}

\usepackage{times}
\usepackage{latexsym}

\usepackage[T1]{fontenc}

\usepackage[utf8]{inputenc}

\usepackage{microtype}

\usepackage{inconsolata}

\usepackage{graphicx}

\usepackage{enumitem}
\usepackage{booktabs}
\usepackage{arydshln} 
\usepackage{listings} 
\newcommand{\modify}[1]{\textcolor{black}{#1}}

\title{Post Persona Alignment for Multi-Session Dialogue Generation}

\author{
 \textbf{Yi-Pei Chen\textsuperscript{1,2}}\quad
 \textbf{Noriki Nishida\textsuperscript{1}}\quad
 \textbf{Hideki Nakayama\textsuperscript{2}}\quad
 \textbf{Yuji Matsumoto\textsuperscript{1}}\quad
\\\\
 \textsuperscript{1}RIKEN AIP\quad
 \textsuperscript{2}The University of Tokyo
}

\begin{document}
\maketitle
\begin{abstract}
Multi-session persona-based dialogue generation presents challenges in maintaining long-term consistency and generating diverse, personalized responses. While large language models (LLMs) excel in single-session dialogues, they struggle to preserve persona fidelity and conversational coherence across extended interactions. Existing methods typically retrieve persona information before response generation, which can constrain diversity and result in generic outputs. We propose Post Persona Alignment (PPA), a novel two-stage framework that reverses this process. PPA first generates a general response based solely on dialogue context, then retrieves relevant persona memories using the response as a query, and finally refines the response to align with the speaker's persona. This post-hoc alignment strategy promotes naturalness and diversity while preserving consistency and personalization. Experiments on multi-session LLM-generated dialogue data demonstrate that PPA significantly outperforms prior approaches in consistency, diversity, and persona relevance, offering a more flexible and effective paradigm for long-term personalized dialogue generation.
\end{abstract}

\input{1_content}

\section*{Acknowledgments}
This work was supported by JST-CRONOS Grant Number JPMJCS24K8, JSPS KAKENHI Grant Number JP23K28139, and the Institute of AI and Beyond of the University of Tokyo.

\bibliography{3_custom}

\noindent\rule{\columnwidth}{0pt}\newpage
\appendix
\input{2_appendix}

\end{document}

%% file: 1_content.tex
\section{Introduction}

Multi-session persona-based dialogue generation aims at generating responses based on the persona sentences, current dialogue context, and dialogue history sessions \cite{xu-etal-2022-beyond}. 
With the development of large language models (LLMs), current dialogue systems are proficient in single-session dialogue generation, and even able to produce responses more natural than human-authored ones \cite{kim-etal-2023-soda}. Despite their impressive capabilities in generating coherent single-session dialogues, LLMs still face challenges in multi-session dialogue generation.

\begin{figure}
    \centering
    \includegraphics[trim={0.8cm 0cm 0cm 0cm},clip,width=\linewidth]{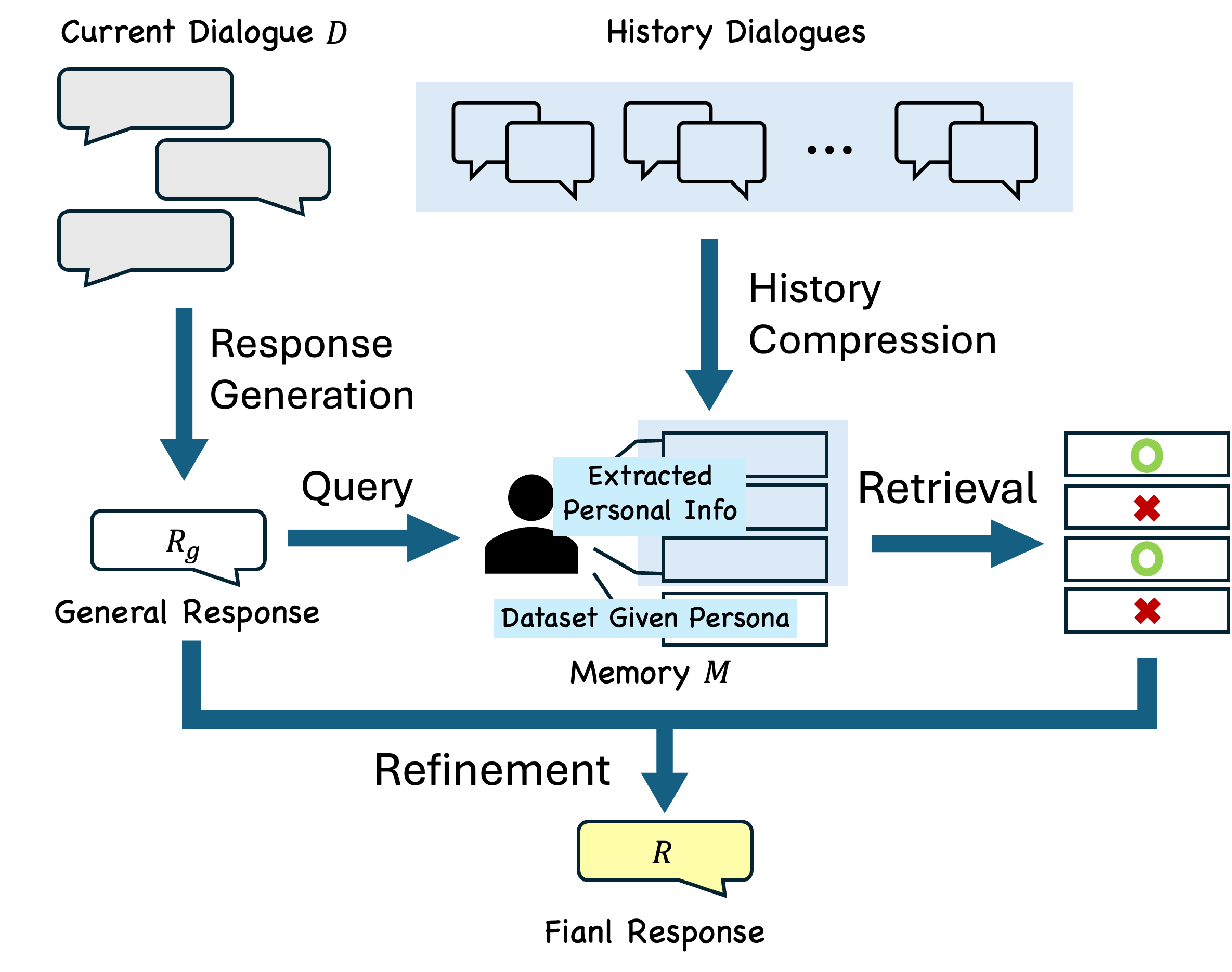}
    \caption{Overview of our PPA framework.}
    \label{fig:overview}
\end{figure}

The first challenge is maintaining consistency as the dialogue history becomes long. 
Research has found that LLM agents cannot adhere to their own persona or be consistent with content in previous conversations after multi-session dialogues \cite{chu2024cohesive}. 
As the dialogue sessions increase, it is necessary to truncate, summarize, or select parts of the history to fit within input token limits \cite{xu-etal-2022-beyond,bae-etal-2022-keep,kwon-etal-2023-ground,xu-etal-2022-long,kim-etal-2024-commonsense}, which might lead to the loss of critical information. 
\modify{While recent LLMs offer greatly extended context windows that can accommodate the full history, prior studies \cite{liu2024lost} show that they may still fail to retrieve and reason over it effectively, struggling to track subtle content shifts and recall information from early interactions.}
The second challenge lies in enhancing the diversity of generated dialogues. 
While less discussed in persona-aware dialogue literature, recent studies have raised concerns about the homogenization and lack of richness in LLM-generated text \cite{chung-etal-2023-increasing,reif2023visualizing,park2024diminished,yu2024large}. 
Retrieval-based multi-session systems often use the last utterance as a query to retrieve relevant persona information. However, this typically yields narrowly scoped content, limiting the opportunity for diverse or novel responses.

To address these challenges, we propose a novel framework called Post Persona Alignment (PPA). 
Unlike conventional approaches that retrieve persona memories before generating responses, PPA reverses the order: it first generates a general response based solely on the dialogue context, then uses that response as a query to retrieve relevant persona memories, and finally refines the response to ensure consistency and personalization. 
The framework is shown in Fig.~\ref{fig:overview}. 
This post-hoc alignment strategy reflects how humans often communicate, i.e., by initially responding naturally to the flow of conversation, then revising or extending their reply based on what they know about the listener. 
Compared to pre-alignment methods, PPA offers more freedom for the model to explore plausible and creative directions before aligning with persona traits, thereby improving both naturalness and diversity.
We evaluate multiple generation strategies with different history representations on LLM-generated multi-session dialogues. 
Our experiments show that PPA significantly improves the consistency, diversity, and personalization of generated responses compared to prior baselines.

\section{Related Work}

Most prior work on personalized dialogue generation adopts a pre-alignment strategy, conditioning the response generation on predefined persona profiles from the outset. 
The Persona-Chat dataset \cite{Zhang2018Personalizing} pioneered this setting by incorporating persona sentences into chit-chat models. Subsequent work extended this paradigm by integrating persona embeddings \cite{bak2019persona} or auxiliary classifiers to detect relevant persona usage \cite{qian2018assigning}. 
These approaches improved persona consistency but often led to generic, rigid, or contextually awkward responses when persona facts were injected indiscriminately.
More recent methods recognized this limitation and adopted multi-stage pipelines to balance coherence and persona relevance. For example, \citet{song-etal-2020-generating} proposed deleting and rewriting persona-inconsistent words post-generation. Similarly, SimOAP \cite{zhou-etal-2023-simoap} and BoB \cite{song-etal-2021-bob} use over-sampling and entailment-based filtering to maintain both fluency and consistency. These trends highlight the growing interest in \textit{decoupling} persona alignment from initial response generation, a direction we pursue further.

Retrieval-augmented dialogue generation has proven effective for incorporating long-term or external knowledge into conversations. Wizard-of-Wikipedia \cite{dinanwizard} retrieves factual background to ground responses in real-world knowledge. Similarly, persona facts can be viewed as structured knowledge: \citet{ma2021one} and \citet{zheng2020pretraining} proposed models that separately encode profile memory and interaction history, improving personalization. 
However, memory retrieval is often driven by the current dialogue context, which limits flexibility and may omit relevant long-range cues. 
To improve coherence and control input length, recent works propose summarizing histories \cite{xu-etal-2022-beyond} or retrieving salient utterances or tokens \cite{xu-etal-2022-long,zhong-etal-2022-less}, though such compression can sacrifice nuanced personal information. Rule-based personal knowledge extraction has emerged as a more effective way to preserve speaker-specific details \cite{mousavi-etal-2023-response}.

Maintaining consistency over multiple sessions is particularly challenging due to memory constraints and the risk of context drift. Hierarchical models such as HRED \cite{serban2016building} were early attempts to encode multi-turn coherence, while modern LLM-based systems simply concatenate long histories, leading to token budget issues and degraded performance \cite{liu2024lost}. 
Multi-Session Chat (MSC) \cite{xu-etal-2022-beyond} has sparked research into long-term memory design \cite{kwon-etal-2023-ground,kim-etal-2024-commonsense}. LLM-based multi-session dialogues further increase dataset size \cite{jang-etal-2023-conversation}, with \citet{Park2023GenerativeAgents} providing some of the longest available sessions. 

In contrast to prior systems that integrate persona information before generation, our proposed framework delays persona grounding until after an initial response is generated. The response is first generated from the dialogue context, then used to retrieve relevant persona memories, which are finally incorporated through refinement. This reversal of the retrieval-generation order yields more natural, diverse responses without sacrificing persona consistency. By grounding after generation, PPA enables targeted alignment with the model's intended output, avoiding unnecessary or rigid persona mentions. This flexibility sets PPA apart from previous persona-grounded or retrieval-augmented methods, offering an effective solution for multi-session dialogue with consistent personalization.

\section{Post Persona Alignment Framework}

We propose a two-stage generation framework called Post Persona Alignment (PPA), which defers persona grounding until after a general response is generated. This design contrasts with conventional persona-aware generation that aligns responses to personas \textit{before} generation begins. Instead, our method first produces a preliminary response guided solely by dialogue context, then selectively retrieves relevant persona information based on that response, and finally refines the output for personalization and consistency. Figure~\ref{fig:overview} illustrates the overall architecture.

\subsection{Stage 1: History Compression via Personal Knowledge Extraction}
\label{sec:persona_augmentation}
To ensure relevant information from multi-session dialogue history is retained within input limits, we compress history into structured personal knowledge. Instead of retrieving raw utterances or summaries, we extract a set of (name, relation, object) triples that represent salient personal facts from each prior session.
These triples are obtained via prompt-based extraction using an LLM, then verbalized into natural language sentences and stored in a memory module alongside any predefined personas. This process helps isolate stable, identity-relevant attributes from conversational noise.
We found this form of compression particularly effective. Raw utterances often contain ambiguous pronouns or speaker tags (e.g., \texttt{Speaker A: "He said you like hiking"}), which can confuse models during retrieval. The explicit structuring of information into memory improves both retrieval precision and interpretability.

\subsection{Stage 2: Response-First Retrieval and Refinement}
Unlike traditional methods that retrieve persona memories based on the dialogue context, PPA flips this order.

\paragraph{Step 1. Response Generation:} 
Given only the current dialogue context $D$, we first generate a general response $R_g$ without conditioning on any persona information. This allows the model to freely express what it deems an appropriate and natural reply in the moment. 
The prompt for $R_g$ generation is shown in Appendix~\ref{sec:prompt_gen}.

\paragraph{Step 2. Response-Guided Memory Retrieval:} 
Next, we use $R_g$ as the retrieval query to compute its similarity with the memory pool (consisting of persona facts and extracted history). We retrieve the top-$k$ most relevant entries $M_k$ whose similarity is higher than the retrieval threshold $\theta$.


\paragraph{Step 3. Response Refinement:} 
Finally, we refine the initial response $R_g$ using both the retrieved persona memories $M_k$ and the original dialogue context $D$, producing a final response $R$ that is consistent with both the conversation flow and the speaker’s persona.
The prompt for generating the final response is shown in Appendix~\ref{sec:prompt_genrefine}.

This post-hoc refinement encourages more diverse and context-appropriate responses while enforcing personalization \textit{after} the model has already formulated its communicative intent. It mirrors how humans often revise their replies—adding relevant details after they’ve begun to speak.

\section{Experiment}

\subsection{Setup}
We evaluate the effectiveness of our proposed PPA framework against several baseline methods on multi-session persona-based dialogue generation.
We compare PPA with the following strategies \modify{to reflect representative prior approaches in persona-based dialogue literature}:
\textbf{DirectGen} \cite{lee-etal-2022-personachatgen, chen-etal-2023-places, Park2023GenerativeAgents} directly provides the persona, history, and current dialogue as a single input to the generator.
\textbf{DialogRetr} \cite{ma2021one, xu-etal-2022-long, kim-etal-2024-commonsense} retrieves top-$k$ relevant persona/history entries based on the dialogue context, and feeds them to the generator.
\textbf{SimOAP} \cite{zhou-etal-2023-simoap} \modify{represents multi-stage response selection approach. They} generate lots of responses from the dialogue and selects the best one based on coherence and consistency with history and persona.
\textbf{BoB} \cite{song-etal-2021-bob} \modify{is a well-known refinement-based approach, where} a neural model takes persona and dialogue as input and learns to refine the output response implicitly in latent space.

We use LLM-generated multi-session dialogues from \citet{Park2023GenerativeAgents}, which have some of the longest dialogue histories available (average 11.9 sessions per dialogue). 
We use GPT-3.5 for both response generation and refinement in PPA. 
The retriever computes cosine similarity between response embeddings and persona memory using SentenceBERT \cite{reimers-2019-sentence-bert}. We set the retrieval threshold $\theta=0.2$ and top-$k=5$ for all methods. 
\modify{Note that our goal is to compare generation paradigms -- post-hoc persona alignment (PPA) versus conventional pre-conditioning or early retrieval approaches -- under equal model and retrieval settings, independent of backbone or embedding choice. }

Following prior works, we evaluate:
\begin{itemize}[topsep=1pt,itemsep=2pt,leftmargin=12pt]
    \item \textbf{Consistency (C-Score)}: whether the response aligns with persona and historical information \cite{madotto-etal-2019-personalizing, cao-etal-2022-model}.
    \item \textbf{Personalization (Persona-F1)}: overlap of generated content with known persona facts \cite{ijcai2019p706}.
    \item \textbf{Diversity (Entropy)}: n-gram entropy to measure lexical richness \cite{shannon1951prediction, zhang2018generating}.
    \item \textbf{ROUGE}: lexical overlap with the reference response \cite{lin-2004-rouge}, though we acknowledge that similarity-based metrics are limited in dialogue settings \cite{liu-etal-2016-evaluate, yeh-etal-2021-comprehensive}.
\end{itemize}

\subsection{Results and Discussion}

\paragraph{How Does PPA Compare to Existing Strategies?}
Table~\ref{tab:result_strategy} presents the main results across all strategies. The proposed PPA achieves the highest scores in consistency (C-Score), diversity (Entropy), and personalization (P-F1), outperforming all baselines by a significant margin.

Among the baselines, DirectGen performs surprisingly well, benefiting from full access to dialogue context, persona, and history. However, its performance in personalization is limited by the entangled input structure, which often fails to prioritize the most relevant persona information. 
DialogRetr, which retrieves memory based on context, struggles to retrieve varied or nuanced persona information, leading to the lowest overall scores. 
SimOAP, while designed to enhance diversity and coherence, achieves the best P-F1 among baselines but falls short of PPA in both consistency and entropy. 
BoB, despite its design for persona refinement, produces extremely short and generic outputs, yielding poor performance across all metrics. 

\begin{table}[t]
\centering
\resizebox{1.0\linewidth}{!}{
\begin{tabular}{lcccc}
    \toprule
    \textbf{Strategy} & \textbf{C-Score} & \textbf{ENTR} & \textbf{P-F1} & \textbf{ROUGE} \\
    \midrule
    BoB & 0.018 & 2.87 & 0.017 & 0.073 \\
    DirectGen & 0.221 & 5.18 & 0.092 & \textbf{0.241} \\
    DialogRetr & 0.182 & 5.03 & 0.081 & 0.210 \\
    SimOAP & 0.206 & 5.07 & 0.100 & 0.228 \\
    PPA (ours) & \textbf{0.456} & \textbf{5.75} & \textbf{0.146} & 0.182 \\
    \bottomrule
\end{tabular}
}
\caption{Results of response generation strategies.}
\vspace{-2mm}
\label{tab:result_strategy}
\end{table}

\begin{table}[!ht]
\centering
\resizebox{1.0\linewidth}{!}{
\begin{tabular}{lcccc}
    \toprule
    \textbf{Query Type} & \textbf{C-Score} & \textbf{ENTR} & \textbf{P-F1} & \textbf{ROUGE} \\
    \midrule
    Context & 0.363 & 5.566 & 0.134 & 0.205 \\
    Response ($R_g$) & 0.456 & \textbf{5.751} & 0.146 & 0.182 \\
    Gold & \textbf{0.554} & 5.671 & \textbf{0.147} & \textbf{0.214} \\
    \bottomrule
\end{tabular}
}
\caption{Retrieval query comparison in PPA.}
\label{tab:result_query}
\end{table}

\begin{table}[!ht]
\centering
\resizebox{\linewidth}{!}{
    \begin{tabular}{lcccc}
    \toprule
    \textbf{History Type} & \textbf{C-Score} & \textbf{ENTR} & \textbf{P-F1} & \textbf{ROUGE} \\
    \midrule
    Utterance & 0.359 & 5.510 & 0.100 & \textbf{0.202} \\
    Summary & 0.406 & 5.688 & 0.135 & 0.188 \\
    Persona & \textbf{0.456} & \textbf{5.751} & \textbf{0.146} & 0.182 \\
    \bottomrule
    \end{tabular}
}
\caption{PPA performance with different history representations.}
\vspace{-0.5em}
\label{tab:result_history}
\end{table}

In contrast, our proposed PPA framework consistently balances between coherence, diversity, and persona grounding by first generating a natural response and then selectively incorporating relevant persona knowledge during refinement. This two-stage approach offers flexibility in generation while enabling precise, post-hoc persona alignment.

\modify{Examining output responses of each strategy, we found that PPA usually incorporates more information from its memory than other strategies. As shown in Appendix~\ref{sec:output_examples}, PPA and DialogRetr contain information mentioend in previous dialogues, while DirectGen and SimOAP generate more general responses.}

\paragraph{Does Response-Guided Retrieval Improve Persona Relevance?}
A core hypothesis of PPA is that retrieving memory \textit{after} generating a general response yields more relevant and diverse persona content. Table~\ref{tab:result_query} compares three query strategies: using the dialogue context (Context), using the generated response ($R_g$), and using the ground-truth response (Gold).
Both Gold and $R_g$ queries outperform Context on all metrics, confirming the effectiveness of our response-guided retrieval approach. Notably, the memory retrieved by response queries often contains a richer variety of persona facts, enabling smoother topic transitions and more personalized outputs.

\paragraph{What Type of History Best Supports Post-hoc Alignment?}
We further test PPA with different forms of compressed history: raw utterances (Utterance), summaries (Summary), and structured personal information (Persona). As shown in Table~\ref{tab:result_history}, using extracted persona knowledge yields the best performance across all metrics. This supports our claim that structured and explicit personal knowledge is more effective than raw or summarized context when performing post-hoc alignment.

\section{Conclusion}
We introduced Post Persona Alignment (PPA), a novel framework for multi-session persona-based dialogue generation that defers persona grounding until after an initial response is generated\modify{, reversing the conventional retrieval-generation order. 
PPA structures information into salient, memory-like facts, and retrieves them only when truly relevant to the model's communicative intent. This approach reduces the cognitive load on the model and improves controllability,} enabling more natural, diverse, and personalized dialogues. Experiments demonstrate that PPA significantly improves consistency, diversity, and persona relevance over prior methods, offering a flexible and effective solution for long-term conversational modeling.

\section*{Limitation}
First, the effectiveness of PPA hinges on the quality of the initial response generated without persona conditioning. If it lacks coherence or relevance to the dialogue context, the subsequent retrieval and refinement stages may reinforce or propagate suboptimal content rather than correct it.
Second, our approach relies on embedding-based similarity between the initial response and persona memory for retrieval. This introduces sensitivity to the quality of the embedding space and the representational alignment between responses and memory entries. In particular, semantically important but lexically dissimilar content may be overlooked by SentenceBERT, limiting retrieval coverage.
Third, as the number of persona facts and extracted knowledge triples increases over many sessions, maintaining and retrieving from an expanding memory pool may introduce computational and latency bottlenecks. Moreover, fine-grained control over long-term memory retention (e.g., forgetting outdated traits or events) remains an open challenge.

%% file: 2_appendix.tex
\section{Prompts}
\label{sec:prompts}

\subsection{Prompt for Response Generation}
\label{sec:prompt_gen}
See Figure~\ref{fig:prompt_gen}.

\begin{figure}[ht]
\centering
\begin{minipage}{0.95\linewidth}
\lstset{
     basicstyle=\ttfamily\small,
     columns=fullflexible,
     keepspaces=true,
     showstringspaces=false,
     breaklines=true,
     breakatwhitespace=true,
     breakindent=0pt,
     frame=single,
     backgroundcolor=\color{gray!5}, 
     captionpos=b
}
\begin{lstlisting}
{speaker} is chatting with {other}.

# The current conversation between {other} and {speaker} is as follows:
{dialogue}

# Task: Output {speaker}'s response to {other} in JSON.
Format: {{speaker}: <response>}
\end{lstlisting}
\caption{Prompt used for response generation. Curly braces \{\} indicate variable substitution.}
\label{fig:prompt_gen}
\end{minipage}
\end{figure}

\subsection{Prompt for Response Refinement} 
\label{sec:prompt_genrefine}
See Figure~\ref{fig:prompt_genrefine}.

\begin{figure}[ht]
\centering
\begin{minipage}{0.95\linewidth}
\lstset{
     basicstyle=\ttfamily\small,
     columns=fullflexible,
     keepspaces=true,
     showstringspaces=false,
     breaklines=true,
     breakatwhitespace=true,
     breakindent=0pt,
     frame=single,
     backgroundcolor=\color{gray!5}, 
     captionpos=b
}
\begin{lstlisting}
{speaker} is chatting with {other}. 
Their conversation is as follows: 
{dialogue}

{speaker} was about to reply: "{response}"

# Task: Refine {speaker}'s response with the following information:
{memory}

# Output {speaker}'s response to {other} in JSON.
Format: {{speaker}: <{speaker}'s refined response>}\\
\end{lstlisting}
\caption{Prompt used for response refinement.}
\label{fig:prompt_genrefine}
\end{minipage}
\end{figure}

\section{Example of Different Strategy Outputs}
\label{sec:output_examples}
See Table~\ref{tab:output_examples}.
\input{tables/output_examples}

%% file: tables/output_examples.tex
\begin{table*}
\begin{tabular}{lp{0.82\linewidth}}
\toprule
\multicolumn{2}{l}{Current Dialogue} \\
\cmidrule(lr){1-2}
Francisco& Hey Rajiv! How's the guitar practice going?\\
Rajiv& It's going alright. I'm still learning the basics.\\
Francisco& That's cool. Have you thought about incorporating your guitar playing into your artwork?\\
Rajiv& Actually, I have. I was thinking about using sound waves as a way to create mathematical patterns.\\
Francisco& That sounds really interesting. We should definitely collaborate on a project involving artwork and music.\\
Rajiv& Definitely. We could also explore incorporating poetry into the mix.\\
Francisco& I like the way you think. Speaking of collaborations, have you signed up for those improv classes yet? \\
\midrule
\multicolumn{2}{l}{Selected Utterance from History Dialogues} \\
\cmidrule(lr){1-2}
Latoya & Have you ever considered attending an improv class? \\
Rajiv& Actually, I have. Hailey Johnson invited me on her podcast to talk about my guitar playing, and she also mentioned attending an improv class together.\\
\hdashline
Francisco & That's awesome. I'm planning to potentially collaborate with Abigail Chen and Jennifer on artistic projects involving poetry and artwork with mathematical patterns. \\
\midrule
Strategy & Response of Rajiv \\
\cmidrule(lr){1-2}
BoB & Yes, I'm just trying to get a job with my new job. \\

DirectGen& Yes, I'm planning to attend improv classes soon. I think it would be a great way to explore connections between art, creativity, and spontaneity. \\

DialogRetr& I haven't signed up for the improv classes yet, but I'm definitely interested. \textbf{Let's plan to attend them together with Hailey Johnson!} \\

SimOAP & I have been considering taking improv classes. It could be a fun and creative way to explore new ideas and collaborate with others. \\

PPA & Not yet, but I'm definitely considering it. Improv could really help with thinking on my feet during creative projects. \textbf{I'm also interested in exploring the connections between math, nature, and art, especially after hearing about your collaborations with Abigail Chen and Jennifer on artistic projects involving poetry and artwork with mathematical patterns. It would be great to bring Hailey Johnson along to the improv classes too. }\\
\bottomrule
\end{tabular}
\caption{Example outputs. PPA provides the richest information from dialogue history (marked in bold), making the response more diverse and personalized.}
\label{tab:output_examples}
\end{table*}